\documentclass[10pt,twocolumn,letterpaper]{article}

\usepackage[pagenumbers]{cvpr} 
\usepackage{graphicx}
\usepackage{amsmath}
\usepackage{amssymb}
\usepackage{booktabs}
\usepackage{multicol}
\usepackage{multirow}
\usepackage[accsupp]{axessibility}
\usepackage[pagebackref,breaklinks,colorlinks]{hyperref}

\usepackage[capitalize]{cleveref}
\crefname{section}{Sec.}{Secs.}
\Crefname{section}{Section}{Sections}
\Crefname{table}{Table}{Tables}
\crefname{table}{Tab.}{Tabs.}

\begin{document}

\title{Unified Keypoint-based Action Recognition Framework \\ via Structured Keypoint Pooling}

\author{
Ryo Hachiuma\thanks{~Equal contribution.}, ~~Fumiaki Sato\footnotemark[1], ~~Taiki Sekii \\
Konica Minolta, Inc. \\
{\tt\small \{rhachiuma,fumiaki.sato.jp,taiki.sekii\}@gmail.com} \\
}

\maketitle

\begin{abstract}
This paper simultaneously addresses three limitations associated with conventional skeleton-based action recognition; skeleton detection and tracking errors, poor variety of the targeted actions, as well as person-wise and frame-wise action recognition.
A point cloud deep-learning paradigm is introduced to the action recognition, and a unified framework along with a novel deep neural network architecture called {\rm Structured Keypoint Pooling} is proposed.
The proposed method sparsely aggregates keypoint features in a cascaded manner based on prior knowledge of the data structure (which is inherent in skeletons), such as the instances and frames to which each keypoint belongs, and achieves robustness against input errors.
Its less constrained and tracking-free architecture enables time-series keypoints consisting of human skeletons and nonhuman object contours to be efficiently treated as an input 3D point cloud and extends the variety of the targeted action.
Furthermore, we propose a {\rm Pooling-Switching Trick} inspired by Structured Keypoint Pooling.
This trick switches the pooling kernels between the training and inference phases to detect person-wise and frame-wise actions in a weakly supervised manner using only video-level action labels.
This trick enables our training scheme to naturally introduce novel data augmentation, which mixes multiple point clouds extracted from different videos.
In the experiments, we comprehensively verify the effectiveness of the proposed method against the limitations, and the method outperforms state-of-the-art skeleton-based action recognition and spatio-temporal action localization methods.
\end{abstract}

\vspace{-5mm}
\section{Introduction}
\label{sec:intro}
Recognizing the actions of a person in a video plays an essential role in various applications such as robotics~\cite{Rodomagoulakis2016ICASSP,Lee2019IROS} and surveillance cameras~\cite{Su2020ECCV,Cheng2021ICPR,Islam2021IJCNN}.
The approach to the action recognition task differs depending on whether leveraging appearance information in a video or human skeletons\footnote{Joints or keypoints specific to a person are referred to as skeletons for clarity, although some are not actual human joints.} detected in the video.
The former appearance-based approaches~\cite{Simonyan2014Neurips,Ng2015CVPR,Tran2015ICCV,Feichtenhofer2016Neurips,Carreira2017CVPR,Tran2018CVPR,Xie2018ECCV,Feichtenhofer2019ICCV,Girdhar2019CVPR,Feichtenhofer2020CVPR,Duan2020ECCV,Islam2021IJCNN,Cheng2021ICPR,Arnab2021ICCV,Liu2022CVPR} directly use video as an input to deep neural networks (DNNs) and thus even can recognize actions with relatively small movements.
However, they are less robust to appearances of the people or scenes that differ from the training data~\cite{Weinzaepfel2021IJCV,Moon2021CVPR}.
On the other hand, the latter skeleton-based approaches~\cite{Sijie2018AAAI,Li2019CVPR,Liu2020CVPR,Zhang2020CVPR,Moon2021CVPR,Su2020ECCV,Chen2021ACMMM,Cai2021WACV,Chen2021AAAI,Duan2022CVPR,Chi2022CVPR} are relatively robust to such appearance changes of a scene or a person because they only input low-information keypoints detected using the multi-person pose estimation methods~\cite{Cao2017CVPR,Sekii2018ECCV,Sun2019CVPR}.

Starting from ST-GCN~\cite{Sijie2018AAAI}, various skeleton-based approaches employing graph convolutional networks (GCNs) have emerged~\cite{Shi2019CVPR,Liu2020CVPR,Cai2021WACV,Chen2021AAAI,Chen2021ACMMM,Chi2022CVPR}.
These approaches model the relationship among keypoints by densely connecting them in a spatio-temporal space using GCNs, which treat every keypoint as a node at each time step.
However, most approaches exhibit low scalability in practical scenarios, and further performance improvement is required since they exhibit three limitations regarding network architectures or their problem settings, as described below.

\noindent \textbf{Skeleton Detection and Tracking Errors.}
Conventional GCN-based methods heavily rely on dense graphs, whose node keypoints are accurately detected and grouped by the same instance.
These methods assume that the DNN features are correctly propagated.
Therefore, if false positives (FPs) or false negatives (FNs) occur during keypoint detection, or if the multi-person pose tracking~\cite{Snower2020CVPR,Rafi2020ECCV} fails, such assumptions no longer hold, and the action recognition accuracy is degraded~\cite{Zhu2019KDD,Duan2022CVPR}.

\noindent \textbf{Poor Variety of the Targeted Actions.}
Conventional approaches limit the number of input skeletons to at most one or two.
Therefore, the recognition of actions performed by many people or those interacting with nonhuman objects is an ill-posed problem.
On the other hand, for a wide range of applications, it is desirable to eliminate such restrictions and target a variety of action categories.

\noindent \textbf{Person-wise and Frame-wise Action Recognition.}
Conventional approaches classify an entire video into actions, while practical scenes are complex and include multiple persons performing different actions in different time windows.
Hence, recognizing each person's action for each frame (\textit{spatio-temporal action localization}) is necessary.

In this paper, a unified action recognition framework and a novel DNN architecture called \textit{Structured Keypoint Pooling}, which enhances the applicability and scalability of the skeleton-based action recognition (see \cref{fig:teaser}), is proposed to simultaneously address the above three limitations.
Unlike previous methods, which concatenate the keypoint coordinates and input them into a DNN designed on a predefined graph structure of a skeleton, the proposed method introduces a point cloud deep-learning paradigm~\cite{Qi2017CVPR,Qi2017Neurips,Zhao2021ICCV} to the action recognition and treats a set of keypoints as an input 3D point cloud.
PointNet~\cite{Qi2017CVPR}, which was proposed in such a paradigm, is an innovative research, whose output is permutation-invariant to the order of the input points.
It extracts the features for each input point and sparsely aggregates them to the output feature vector using \textit{Max-Pooling}.
Unlike PointNet, the proposed network architecture aggregates the features extracted from the point cloud in a cascaded manner based on prior knowledge of the data structure, which is inherent in the point cloud, such as the frames or the detection results of the persons (instances) to which each keypoint belongs.
As a result, it is less constrained than conventional approaches and tracking-free. Also, its feature propagation among keypoints is relatively sparse. Therefore, the range of the DNNs affected by the keypoint errors (\eg, FPs, FNs, and tracking errors) associated with the first robustness limitation can also be limited.

In addition, the permutation-invariant property of the input in the proposed network architecture eliminates the constraints of the data structure and size (\eg, number of instances and pose tracking) found in the GCN-based methods.
This property is exploited, and the {\it nonhuman object} keypoints\footnote{Nonhuman objects are referred to as objects for simplicity.} defined on the contour of the objects are used as an input in addition to human skeletons.
Thus, the second target-action limitation mentioned above is addressed by increasing the input information without relying on the appearances while avoiding overfitting on them~\cite{Choi2019Neurips,Weinzaepfel2021IJCV,Moon2021CVPR}.

Finally, the third multi-action limitation is addressed by extending the proposed network architecture concept to a weakly supervised spatio-temporal action localization, which only requires a video-level action label during training.
This is achieved using the proposed \textit{Pooling-Switching Trick} inspired by Structured Keypoint Pooling, which switches the pooling structures according to the training and inference phases.
Furthermore, this pooling-switching technique naturally enables the proposed training scheme to introduce novel data augmentation, which mixes multiple point clouds extracted from different videos.

In summary, our main contributions are three-fold:
(1) We propose Structured Keypoint Pooling based on point cloud deep-learning in the context of action recognition.
This method incorporates prior knowledge of the data structure to which each keypoint belongs into a DNN architecture as an inductive bias using a simple Max-Pooling operation.
(2) In addition to the human skeletons, object keypoints are introduced as an additional input for skeleton-based action recognition.
(3) A skeleton-based, weakly supervised spatio-temporal action localization is achieved by introducing a Pooling-Switching Trick, which exploits the feature aggregation scheme of Structured Keypoint Pooling.

\section{Related Work}
\subsection{Action Recognition}
\noindent \textbf{Appearance-based Action Recognition.}
Numerous prior works rely on RGB images, which are used as inputs to DNNs~\cite{Simonyan2014Neurips,Ng2015CVPR,Tran2015ICCV,Feichtenhofer2016Neurips,Carreira2017CVPR,Tran2018CVPR,Xie2018ECCV,Feichtenhofer2019ICCV,Feichtenhofer2020CVPR,Girdhar2019CVPR}.
In early deep-learning-based approaches, RGB and optical flow images are used as inputs to a 2D convolutional neural network (CNN) to explicitly model the appearance and motion features~\cite{Simonyan2014Neurips,Ng2015CVPR}.
The methods that extract spatio-temporal features using a 3D CNN obtain the motion feature extractors in a data-driven manner~\cite{Tran2015ICCV,Feichtenhofer2016Neurips,Carreira2017CVPR}.
On the other hand, some studies have focused on reducing the computational cost and the number of parameters of a 3D CNN~\cite{Tran2018CVPR,Xie2018ECCV,Feichtenhofer2019ICCV,Feichtenhofer2020CVPR}.
Recently, methods that extract long-range features using the Transformer~\cite{Vaswani2017Neurips} have been proposed~\cite{Girdhar2019CVPR,Arnab2021ICCV,Liu2022CVPR}.
These appearance-based approaches have an advantage over skeleton-based methods because they use more detailed movement features.

\noindent \textbf{Skeleton-based Action Recognition.}
Skeleton-based approaches have been actively investigated since ST-GCN~\cite{Sijie2018AAAI}, which models the relationships among time-series keypoints using GCNs.
Upon the ST-GCN, the robustness and performance of these approaches have been improved by extracting the features from distant keypoints in the spatio-temporal space~\cite{Li2019CVPR,Liu2020CVPR,Chen2021ACMMM,Chen2021AAAI} or by employing efficient graph convolution layers~\cite{Zhang2020CVPR,Cai2021WACV}.
In these methods, the input skeleton sequences can capture only motion information that is immune to contextual nuisances such as background variation and lighting changes~\cite{Choi2019Neurips,Weinzaepfel2021IJCV,Moon2021CVPR}.
Despite their significant success, GCN-based methods exhibit the three limitations mentioned in \cref{sec:intro}.

SPIL~\cite{Su2020ECCV}, which uses an attention mechanism among keypoints, also handles skeleton sequences as an input 3D point cloud and competes with the proposed method only with respect to the network architecture concept.
Unlike SPIL, the proposed method does not rely on such a redundant attention module.
Instead, it introduces a simple and sparse feature aggregation structure, which exploits prior knowledge of the data structure to which each keypoint belongs as an inductive bias.

\subsection{Spatio-temporal Action Localization}
When multiple persons appearing in a video perform different actions in different time windows, according to the third multi-action limitation mentioned in \cref{sec:intro}, this can be handled as a spatio-temporal action localization task.
In the fully-supervised setting, appearance-based approaches~\cite{Yixuan2020ECCV,Pan2021CVPR,Kumar2022CVPR} have been proposed but require dense instance-level annotations during the training.
To reduce the annotation cost, weakly supervised methods~\cite{Cheron2018Neurips,Victor2020CVIU,Anurag2020ECCV} use only a single label for the video as supervision. 
These methods employ the multiple instance learning framework~\cite{Thomas1997AI} for the weakly supervised setting to which this study also focuses on.
Unlike such appearance-based approaches, our input keypoint information is less sensitive to the appearance changes.
In addition, weakly supervised learning is achieved using a simple Pooling-Switching Trick, which exploits our point cloud-based setting and only changes the pooling kernels between the training and inference phases.

\section{Proposed Framework}

\begin{figure*}[tb]
  \centering
  \includegraphics[clip, width=\hsize]{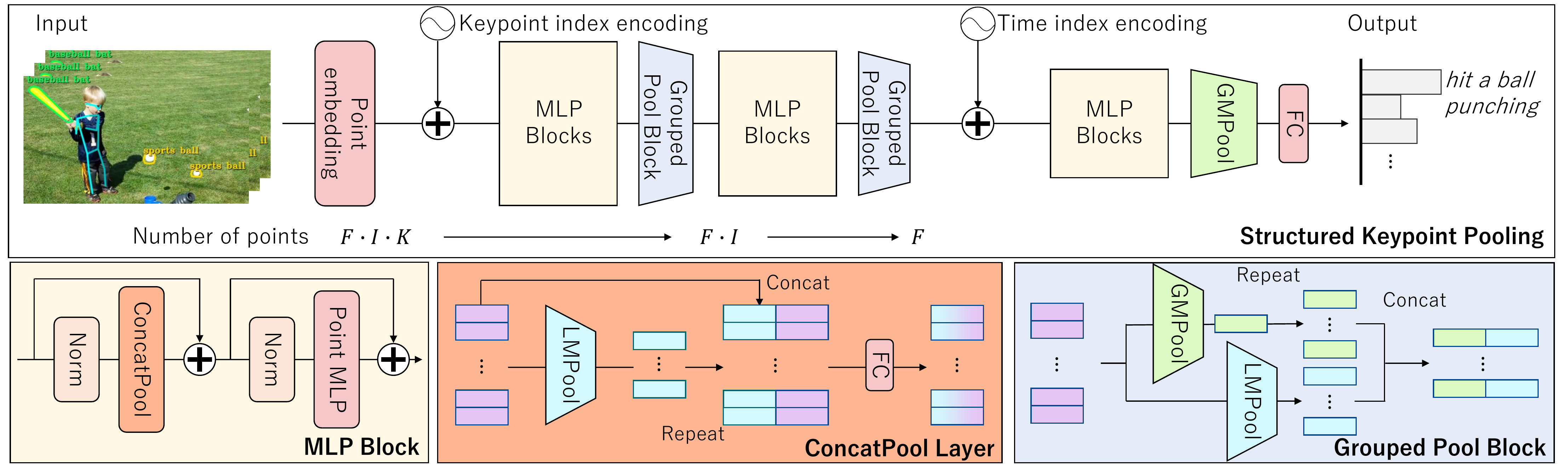}
  \caption{Overview of the Structured Keypoint Pooling network architecture (top) and its original components (bottom).}
  \label{fig:overview}
\end{figure*}

\subsection{Overview}
The proposed network architecture and its components are shown in \cref{fig:overview}.
One of our core ideas is the feature aggregation by a Max-Pooling operator based on groups belonging to the same instance or the same frame (referred to as {\it local groups}).
Limiting the feature-propagation range to the local group is essentially similar to the convolution operation, which extracts the pixel features locally; this is introduced as an inductive bias in our model.
The proposed model essentially consists of only a few conventional DNN modules; nevertheless, its original design and inputs contribute to a significant performance improvement.
In the following, we describe the network architecture along its process and each component in detail.

First, multi-person pose estimation and object keypoint detection are applied to the input video, and human joints as well as object contour points (collectively denoted as keypoints) are obtained.
Then, the keypoints extracted from all frames in the video are treated as a point cloud and used as inputs to the network.
Each keypoint is represented by a four-dimensional vector, which consists of the two-dimensional image coordinates, the confidence score, and the category index of the instance in which the keypoint belongs (\eg, $0$ denotes \textit{person}, $1$ denotes \textit{car}, etc.).
Each element of the input vector is normalized between $0$ and $1$.

Structured Keypoint Pooling~$f_\theta$ predicts logit~$z$, where $z \in \mathbb{R}^C$ for the action recognition task and for the training phase in a weakly supervised action localization task.
As discussed later, $z \in \mathbb{R}^{FI \times C}$ for the inference phase in the action localization task.
$F$ denotes the number of frames in the video clip, $I$ denotes the number of instances per frame, and $C$ denotes the number of target classes.
$\theta$ in $f_\theta$ represents the trainable parameters, and $f_\theta$ mainly consists of MLP Blocks and Grouped Pool Blocks (GPB).
During the training phase, the cross-entropy loss $L_\theta\left(\mathrm{softmax}(z), l\right)$ is computed using the softmax layer and the ground-truth action label $l$; $\theta$ is updated via backpropagation.

The Point embedding layer embeds the input vector into a high-dimensional feature vector using multilayer perceptrons (MLPs).
The weights of the MLP are shared across all keypoints.
We adopt keypoint index encoding, which replaces the position in the original sinusoid positional encoding~\cite{Vaswani2017Neurips} with a keypoint index.
The keypoint index represents its type, for example, $0$ for the \textit{left shoulder} and $1$ for the \textit{right shoulder} regarding the skeleton keypoints; also, it is $0$ for \textit{up left} and $1$ for \textit{up right} regarding these objects.

The MLP Block computes the feature vectors considering the sparse relationships among them via Max-Pooling, and the GPB aggregates such feature vectors into local groups.
Similar to keypoint index encoding, we adopt time index encoding, which encodes the frame index in the video clip.
The feature vectors are finally aggregated by global max-pooling (GMPool) to generate a single feature vector for the entire video.
The logit is predicted via the fully-connected (FC) layers.

The reduction in the number of feature vectors by the GPB is described in the following.
We denote $K$ as the number of keypoints per instance, in addition to $F$ and $I$ defined above.
The number of keypoints input to the network is $F \cdot I \cdot K$, which is reduced to $F \cdot I$ points by the first Grouped Pool Bock that aggregates $K$ keypoint-wise features into a single vector.
Then, the second GPB that aggregates $I$ instance-wise features in a single vector reduces the number of points from $F \cdot I$ to $F$.

The Max-Pooling operator outputs a feature vector by selecting a maximum value for each dimension from $N$ input vectors.
Therefore, elements from maximum $D$ points are selected ($D$ is the feature dimension size of input vectors).
As $N \gg D$ (\eg, $N = F \cdot I \cdot K = 300 \cdot 2 \cdot 18$, $D = 512$) for the skeleton-based action recognition task, most points will be disregarded for the GMPool (pooling across all input points).
Reducing the number of points ($N$) by cascaded feature aggregation and limiting the pooling range using local max-pooling (LMPool), which applies Max-Pooling to each local group to which the input feature vectors belong, are helpful to generate informative and robust feature vectors. The effect of using this cascaded reduction during the feature extraction will be quantitatively verified in \cref{sec:ablation}.

The process in each block is invariant to the position and order of the input feature vectors, and the entire network can handle permutation-invariant inputs.

\subsection{Grouped Pool Block (GPB)}
The GPB consists of GMPool $\phi_G$ and LMPool $\phi_L$.
The first GPB outputs feature vectors containing the number of instances in the video, and the subsequent GPB outputs feature vectors containing the number of frames.

The GPB can be expressed as follows:
\begin{equation}
  Y = \left\{\left[\phi_L\left(X\right)_j; \phi_G\left(X\right)\right]\right\}_{j \in \left\{1,\ldots,M\right\}}.
  \label{eq:gpb}
\end{equation}
$X$ and $Y$ are the matrices of the input and output feature vectors, respectively, as described below.
$M$ denotes the number of local groups in $X$; $M=F \cdot I$ in the first block; $M=F$ in the second block.
Therefore, the input feature vector $x_i \in X$ is grouped into $M$ local groups, and $X$ can be expressed by a concatenated matrix as follows:
\begin{equation}
  X = {\left(x_1,\ldots,x_N\right)}^T = {\left(X_1;\ldots;X_M\right)}^T.
\end{equation}
Consequently, $Y$ is computed using the output vector $y_j$ as follows:
\begin{equation}
Y = {\left(y_1,\ldots,y_M\right)}^T.
\end{equation}
In \cref{eq:gpb}, we concatenate each feature vector $\phi_L\left(X\right)_j$ computed for the local group $j$ and the global feature vector $\phi_G\left(X\right)$ in a channel dimension.
Also, LMPool~$\phi_L(\cdot)$ can be expressed as follows:
\begin{equation}
  \begin{split}
    \phi_L\left(X\right) = \left\{\mathrm{MaxPool}(X_j)\right\}_{j = \left\{1,\ldots,M\right\}},
  \end{split}
\end{equation}
where $\mathrm{MaxPool}(\cdot)$ is the operation used to obtain the max value for each channel from the feature vectors.
GMPool~$\phi_G(\cdot)$ is expressed as follows:
\begin{equation}
  \phi_G(X) = \mathrm{MaxPool}(X).
\end{equation}

\subsection{MLP Block}
The MLP Block consists of two residual blocks.
The first block models the relationship among feature vectors within each local group.
The subsequent block applies MLPs for each feature vector.
Each MLP block is repeated $r$ times.

The first residual block can be written using the input and output matrices $X$ and $Y$, respectively, as follows:
\begin{equation}
  Y = \mathrm{ConcatPool}(\mathrm{Norm}(X)) + X,
\end{equation}
where $\mathrm{Norm(\cdot)}$ is the normalization layer, and $\mathrm{ConcatPool(\cdot)}$ is the learnable layer represented as
\begin{equation}
  \mathrm{ConcatPool}(X) = \left\{\sigma \left(\left[x_i; \phi_L\left(X\right)_{j_i}\right]W_1\right)\right\}_{i \in \left\{1,\ldots,N\right\}},
\end{equation}
where $\sigma(\cdot)$ is a nonlinear activation function and $j_i \in \left\{1,\ldots,M\right\}$ is the local group index of the $i$-th feature vector.
$W_1 \in \mathbb{R}^{2D \times D}$ is a learnable weight matrix, and $D$ is the number of channels of $X$.

The second residual block can be expressed as follows:
\begin{equation}
  Y = \sigma\left(\mathrm{Norm}\left(X\right)W_2\right)W_3 + X,
\end{equation}
where $W_2 \in \mathbb{R}^{D \times \alpha D} $ and $W_3 \in \mathbb{R}^{\alpha D \times D}$ are learnable weight matrices, and $\alpha$ is the MLP expansion ratio.

\subsection{Pooling-Switching Trick for Spatio-Temporal Action Localization}
\begin{figure}[tb]
  \centering
  \includegraphics[clip, width=\hsize]{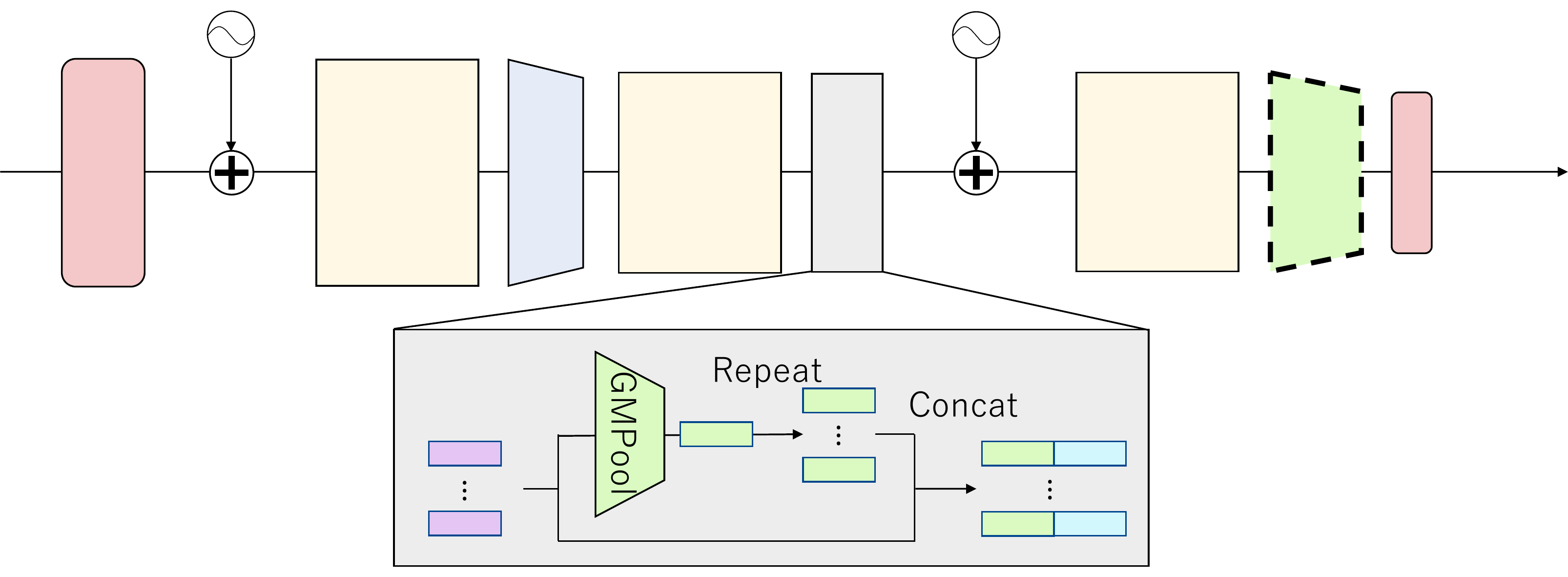}
  \caption{Pooling-Switching Trick for point cloud-based spatio-temporal action localization.
  The modules same as those in \cref{fig:overview} are abbreviated with the same color.
  The dotted GMPool layer is only applied during the training phase.}
  \label{fig:actloc}
\end{figure}

The proposed network architecture of the spatio-temporal action localization is shown in \cref{fig:actloc}.
To avoid aggregating instance-level features into frame-level features, the second GPB in \cref{fig:overview} is changed.
We propose a Pooling-Switching Trick, which switches the group of the pooling (kernel) from the training to the inference phases.
This trick naturally enables our weakly supervised training scheme to introduce the proposed batch-mixing data augmentation.

\noindent \textbf{Weakly Supervised Training.}
During the training, the loss is computed between the ground-truth action label assigned to the input video and the video-level logit predicted by aggregating instance-level features using the last GMPool.
During the inference, the proposed method estimates the actions against targets different from the training, such as each instance, each frame, or each video, by switching the pooling kernel (target local group) at the last GMPool operation.
For the spatio-temporal action localization task, the GMPool operation is simply removed from the network architecture (\cref{fig:actloc}) to estimate the instance-level logit.
The weights of the FC layer are shared across all targets.

\noindent \textbf{Batch-Mixing Augmentation.}
To improve the localization robustness, we propose a novel data augmentation technique in the Pooling-Switching Trick. This technique mixes the point clouds extracted from different videos and promotes classifying multiple actions.
Let $X \in \mathbb{R}^{FI \times D}$ and $l$ denote instance-level features and the corresponding ground-truth one-hot label, respectively.
Two training samples $(X^a, l^a)$ and $(X^b, l^b)$ are mixed for augmentation.

First, we mask two training samples as follows:
\begin{equation}
\hat{X}^a = B \odot X^a, \hat{X}^b = (1-B) \odot X^b,
\end{equation}
where $B \in \mathbb{R}^{FI \times D}$ denotes a binary mask indicating which keypoint is used in the two samples.
Each column vector in $B$ is $0$ or $1$, and $\odot$ denotes the element-wise multiplication.
Also, the ground-truth label is mixed with a certain ratio $\lambda$ as follows:
\begin{equation}
\hat{l} = \lambda l^a + (1-\lambda)l^b.
\end{equation}
A random sampling of the mixing ratio $\lambda$ and the binary mask is followed to the CutMix strategy~\cite{Yun2019ICCV}.

Instead of aggregating a set of feature vectors in the global feature using GMPool \textit{within} each training sample (intra-sample), GMPool (two green boxes in \cref{fig:actloc}) aggregates \textit{between} two training samples (inter-samples) to the global feature vector during the training phase as follows:
\begin{equation}
  \phi_G\left(\hat{X}^a, \hat{X}^b\right) = \mathrm{MaxPool}(\hat{X}^a; \hat{X}^b).
\end{equation}
Finally, the mixed logit $\hat{z}$ is predicted, and the cross-entropy loss $L_\theta\left(\mathrm{softmax}\left(\hat{z}\right), \hat{l} \right)$ is computed.

\section{Experiments}

\subsection{Datasets}

\noindent \textbf{Kinetics-400.} 
The Kinetics-400~\cite{Carreira2017CVPR} dataset is a large-scale video dataset collected from YouTube videos with 400 action classes.
It contains 250K training and 19K validation 10-second video clips.

\noindent \textbf{UCF101 and HMDB51.} 
The UCF101~\cite{Soomro2012Arxiv} and HMDB51~\cite{Kuehne2011ICCV} datasets contain 13K YouTube videos with 101 action labels and 6.7K videos with 51 action labels, respectively.
We employ \textit{split1} for training and test data splitting, according to the previous work~\cite{Duan2022CVPR}.

\noindent \textbf{RWF-2000, Hockey-Fight, Crowd Violence, and Movies-Fight.} 
The RWF-2000~\cite{Cheng2021ICPR}, Hockey-Fight~\cite{CAIP2011Nievas}, Crowd Violence~\cite{CVPR2012Hassner}, and Movies-Fight~\cite{nievas2011violence} datasets are violence recognition datasets.
These datasets contain two types of actions, violence and non-violence, with various people and backgrounds.

\noindent \textbf{Mimetics.} 
The Mimetics dataset~\cite{Weinzaepfel2021IJCV} contains 713 YouTube video clips of mimed actions that form a subset of 50 classes obtained from the Kinetics-400 dataset.
This dataset evaluates human actions with out-of-context appearances different from the Kinetics-400 dataset, and thus the methods have been trained on only the Kinetics-400 dataset.

\noindent \textbf{Mixamo.} 
The Mixamo dataset~\cite{Costa2022WACV} is an action recognition dataset that was proposed for the evaluation of domain adaptation tasks.
This dataset is synthetically generated using the Mixamo library~\cite{mixamo}.
The 3D virtual avatars perform 14 different actions with various backgrounds and objects.
The dataset contains 24K 2D-rendered videos.

\noindent \textbf{UCF101-24.} 
The UCF101-24 dataset~\cite{Soomro2012Arxiv} is a subset of the UCF101 dataset.
Its 24 class action labels are annotated for each bounding box in the videos.
Following the standard practice~\cite{Cheron2018Neurips,Anurag2020ECCV}, we use the corrected annotation~\cite{Singh2017ICCV}.

\subsection{Evaluation Metrics}
We employ Top-1 Accuracy ($\%$) (simply referred to as {\it accuracy}) as the evaluation metric for an action recognition task.
For a spatio-temporal action localization task, we employ Video Average Precision (Video AP) ($\%$) with different 3D IoUs ($0.2$ and $0.5$) as the evaluation metrics.
We use a machine equipped with Intel i7-10700K CPU, 32GB RAM, and GeForce RTX 3080Ti GPU to compute the speed.
See the supplementary material for the implementation details, the hyperparameters of the training, and data augmentations pertaining to all experiments.

\subsubsection{Keypoint Detectors}
\label{sec:keypoint_estimator}

\begin{figure}[tb]
  \centering
  \includegraphics[clip, width=0.95\hsize]{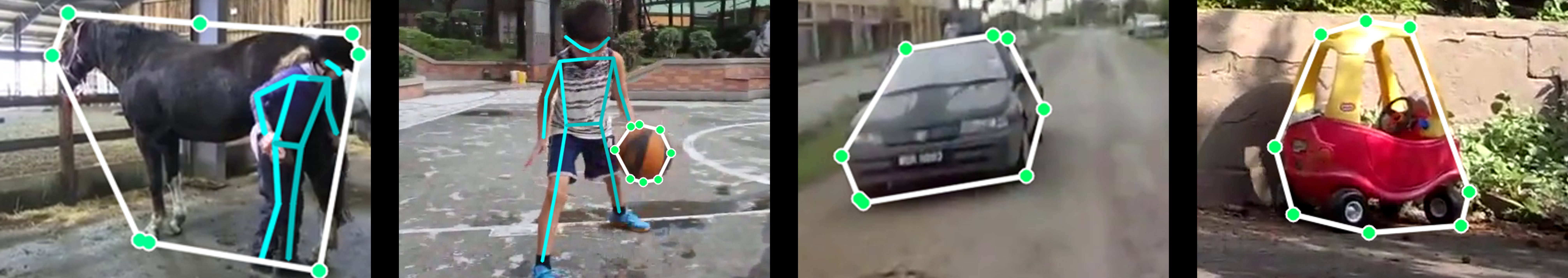}
  \caption{Examples of human skeletons (blue) and eight object contour keypoints (green).}
  \label{fig:ppnv2}
\end{figure}

\noindent \textbf{PPNv2.} The pose proposal networks (PPNv2)~\cite{Sekii2018ECCV, PPNv2} simultaneously detect human skeletons and object keypoints located onto the object contours from an RGB image at a high speed.
They are employed to generate keypoints in an experiment using object information as an input and consist of a Pelee backbone~\cite{Wang2018Neurips} trained on the MS-COCO dataset~\cite{Lin2014ECCV} with both human and object keypoint annotations.
The definition of a human skeleton is the same as the OpenPose~\cite{Cao2017CVPR} definition.
The object keypoints are defined as the eight extreme points on the contours with respect to the eight directions centered on the object (see \cref{fig:ppnv2}).
The input image is resized by $320 \times 224$~$\text{px}^2$.

\noindent \textbf{HRNet.} For a fair comparison with conventional skeleton-based approaches~\cite{Su2020ECCV,Liu2020CVPR,Moon2021CVPR,Duan2022CVPR}, the HRNet~\cite{Sun2019CVPR} is also employed as the human keypoint detector.
The HRNet is a Top-Down pose detector that achieves superior human pose estimation performance.
However, its computational cost, which includes a human detector (Faster R-CNN~\cite{Ren2015Neurips}), is expensive.
We use publicly available HRNet skeletons~\cite{Duan2022CVPR} for the Kinetics-400, UCF101, and HMDB51 datasets.
With the same setting~\cite{Duan2022CVPR}, HRNet skeletons are generated for the RWF-2000, Hockey-Fight, Crowd Violence, Movies-Fight, and Mimetics datasets.

\subsection{Skeleton-based Action Recognition Performance Comparisons on the Kinetics-400}
In \cref{tab:kineticsAcc}, the action recognition accuracy and the speed between the proposed method and conventional skeleton-based approaches are compared on the Kinetics-400 dataset.
It can be observed that the proposed method (Ours w/ objects), which inputs both the skeleton and object keypoints detected by PPNv2, outperforms the conventional methods.
Moreover, its accuracy is improved by $9.2$ percentage-point by introducing object keypoints in addition to the skeletons (Ours w/o objects vs. Ours w/ objects).
The qualitative results are shown in \cref{fig:teaser} (top).

Compared with conventional methods that employ HRNet keypoints~\cite{Sun2019CVPR}, the proposed method outperforms state-of-the-art (SoTA) methods~\cite{Liu2020CVPR,Duan2022CVPR} (MS-G3D and PoseConv3D), while its runtime is $3$x and $96$x faster, respectively, than the runtime of these methods.
Considering ablation studies, as discussed later, these results show that the proposed method overcomes both the first robustness and the second target-action limitations, mentioned in \cref{sec:intro}.

\begin{table*}[tb]
    \begin{minipage}[l]{0.68\textwidth}
  \centering
  \caption{Speed/Accuracy comparison of SoTA skeleton-based action recognition methods on the Kinetics-400 dataset.
  Column Runtime shows the computation time of only the action recognition model.
  Column Total FPS shows the speed, including keypoint detection and action recognition.
  The \textit{joint-bone two-stream ensemble} framework is employed for a fair comparison with conventional methods~\cite{Shi2019CVPR,Liu2020CVPR,Duan2022CVPR}.
  Additionally, we combine HRNet human joints and PPNv2 object keypoints, and the result is 61.4\% (+11.1 percentage-point by using objects).}
  \vspace{-1em}
  \scalebox{0.8}{
  \begin{tabular}{c|c|cc|cc} \hline
  Method & Acc. (\%) & Keypoint Detector & COCO $\text{AP}_\text{kp}$ (\%) & Runtime (ms) & Total FPS \\ \hline
  ST-GCN~\cite{Sijie2018AAAI} & 30.7 & \multirow{3}{*}{OpenPose~\cite{Cao2017CVPR}} & \multirow{3}{*}{56.3} & 4.0 & 85.4 \\ 
  2s-AGCN~\cite{Shi2019CVPR}  & 36.1 & & & 27.6 & 84.8 \\ 
  MS-G3D~\cite{Liu2020CVPR}   & 38.0 & & & 28.2 & 84.8 \\ \hline
  MS-G3D~\cite{Liu2020CVPR}   & 45.1 & \multirow{3}{*}{HRNet~\cite{Sun2019CVPR}} & \multirow{3}{*}{74.6} & 28.2 & 8.8 \\ 
  PoseConv3D~\cite{Duan2022CVPR}  & 47.7 & & & 960.0 & 8.5 \\ 
  Ours w/o objects                        & 50.3 & & & 9.8 & 8.8 \\ \hline
  Ours w/o objects & 43.1 & \multirow{2}{*}{PPNv2~\cite{Sekii2018ECCV}} & \multirow{2}{*}{36.4} & 9.8 & 1913 \\
  Ours w/ objects & \textbf{52.3} & & & 11.2 & 1896 \\ \hline 
  \end{tabular}
  \label{tab:kineticsAcc}
  }
\end{minipage}
\begin{minipage}[r]{0.3\textwidth}
  \hspace{3em}
  \caption{Ablation study of the GPB on the Kinetics-400 dataset with HRNet skeletons.}
  \scalebox{0.8}{
  \begin{tabular}{cc|cc} \hline
      Inst. & Frame & Acc. (\%) & Runtime (ms) \\ \hline
    -            & -            & 47.3 & 89.5 \\ 
    $\checkmark$ & -            & \textbf{48.6} & 7.2 \\ \hline 
    $\checkmark$ & $\checkmark$ & 48.5 & \textbf{4.9} \\ \hline
  \end{tabular} \\
  \label{tab:gpb_ablation}
  }
  \caption{Ablation study of the object keypoint input on the Kinetics-400 dataset with PPNv2 keypoints.}
  \scalebox{0.8}{
  \begin{tabular}{ccc|c} \hline
    Category     & Bbox         & Contours  & Acc. (\%)\\ \hline
    -            & -            & -            & 41.2 \\
    $\checkmark$ & $\checkmark$ & -            & 48.6  \\ \hline
    $\checkmark$ & -            & $\checkmark$ & \textbf{49.2}  \\ \hline
  \end{tabular}
  \label{tab:obj_ablation}
  }
\end{minipage}
  \begin{minipage}[l]{0.55\textwidth}
    \caption{Accuracy Comparison on UCF-101 (U), HMDB51 (HM), Mimetics (Mi), RWF-2000 (R), Hockey-Fight (Ho), Crowd Violence (C), and Movies-Fight (MF)  datasets.}
    \vspace{-1em}
    \scalebox{0.8}{
      \begin{tabular}{c|c|cc|ccccc} \hline 
        Method    & Input                       & U        & HM       & Mi    & R & Ho & C & MF      \\  \hline
        I3D~\cite{Carreira2017CVPR}  & \multirow{5}{*}{\begin{tabular}{c}RGB/ \\ Flow\end{tabular}}   & 95.6          & 74.8             &  -    &  83.4  & 93.4 & 83.4 & 95.8          \\  
        Flow Gated~\cite{Cheng2021ICPR}        &    & -             & -                & -     & 87.3 & 98.0 & 88.8 & 97.3          \\ 
        3D ResNext~\cite{Weinzaepfel2021IJCV}       &  & -             & -                &  10.5            & - & - & - & -     \\ 
        SlowOnly~\cite{Feichtenhofer2019ICCV}           &  & 92.8          & 66.0             & -            & -  & - & - & - \\
        OmniSource~\cite{Duan2020ECCV}                  &   & \textbf{98.6} & \textbf{87.0}    & -            & -  & - & - & -   \\  \hline
        SIP-Net~\cite{Weinzaepfel2021IJCV}              &  \multirow{5}{*}{Skeleton}   & -             & -                & 14.2           & -  & - & - & - \\ 
        IntegralAction~\cite{Moon2021CVPR}              &                             & -             & -                & 15.3          & - & - & - & - \\ 
        PoseConv3D~\cite{Duan2022CVPR}                  &                             & 87.0          & 69.7             & -            & -    & - & - & - \\ 
        SPIL~\cite{Su2020ECCV}                          &                             & -             & -                & -      & 89.3 & 96.8 & 94.5 & 98.5  \\ 
        Ours                                            &                             & 87.8          & 70.9             & \textbf{21.2}&  \textbf{93.4} & \textbf{99.5} & \textbf{94.7} & \textbf{99.0} \\ \hline 
        \end{tabular}
        \label{tab:other_datasets}
    }
  \end{minipage}
  \begin{minipage}[c]{0.4\textwidth}
    \caption{Domain shift experiment on the Mixamo dataset for training and the Kinetics-400 dataset for evaluation. Unsupervised (US) and weakly supervised (WS) domain adaptation (DA) methods are employed as a comparison.}
    \vspace{-1em}
    \hspace{1em}
    \scalebox{0.8}{
    \begin{tabular}{cc|c|c} \hline
      Method & DA & Input  & Acc. (\%) \\ \hline
      I3D~\cite{Carreira2017CVPR} & - & RGB  & 11.2 \\ \hline
      T$\rm{A}^3$N~\cite{Chen2019ICCV}   & \multirow{2}{*}{US} & \multirow{2}{*}{RGB} & 10.0 \\
      C$\rm{O}^2$A~\cite{Costa2022WACV}  &  &  & 16.4 \\ \hline 
      T$\rm{A}^3$N~\cite{Chen2019ICCV} & \multirow{2}{*}{WS} & \multirow{2}{*}{RGB} & 19.1 \\
      C$\rm{O}^2$A~\cite{Costa2022WACV} &  & & 20.1 \\ \hline
      \multirow{2}{*}{Ours} & \multirow{2}{*}{-}  & Skeleton  & 27.6 \\ 
       &    & Skeleton+Object & \textbf{28.4} \\ \hline
      \end{tabular}
        \label{tab:dom_shift}
    }
  \end{minipage}
  \begin{minipage}[l]{0.55\textwidth}
    \caption{Ablation study of the overall framework on the Kinetics-400 dataset with HRNet skeletons.}
    \vspace{-1em}
    \scalebox{0.8}{
    \begin{tabular}{c|cccc} \hline
      \multirow{3}{*}{Design of the GPB} & \multicolumn{3}{c}{Design of the MLP Blocks} \\ 
      & \multirow{2}{*}{Only MLPs} & \multirow{2}{*}{\begin{tabular}{c}1st MLP Block  \\ $\rightarrow$ MS-G3D\end{tabular}}  & \multirow{2}{*}{\begin{tabular}{c}Ours  \\ (MLP+ConcatPool) \end{tabular}} \\ 
      & & & \\ \cline{2-4}
      w/o LMPool & 30.3 & 45.5 & 47.3 \\
      Ours (w/ GPB)  & 44.5 & 45.7 & \textbf{48.5} \\ \hline
      \end{tabular}
    \label{tab:detailed_ablation}
    }
    \end{minipage}
    \hspace{2em}
    \begin{minipage}[c]{0.4\textwidth}
    \caption{Comparison with SoTA weakly supervised spatio-temporal action localization methods on the UCF101-24 dataset.} 
    \vspace{-1em}
      \scalebox{0.8}{
    \begin{tabular}{c|c|cc} \hline
      Method & Input & AP@0.2 & AP@0.5 \\ \hline
      Escorcia~\etal~\cite{Victor2020CVIU} & \multirow{3}{*}{RGB} & 45.5 & - \\ 
      Ch\'{e}ron~\etal~\cite{Cheron2018Neurips} &  & 43.9 & 17.7 \\ 
      Anurag~\etal~\cite{Anurag2020ECCV} &  & 61.7 & 35.0 \\ \hline
      Ours w/o Mix. Aug. & \multirow{2}{*}{Skeleton} & 60.4 & 37.4 \\
      Ours w/ Mix. Aug. &   & \textbf{61.8} & \textbf{38.0} \\ \hline
      \end{tabular}
    \label{tab:act_loc}
       }
  \end{minipage} 
\end{table*}

\subsection{Robustness against Skeleton Detection and Tracking Errors}
\begin{figure}[tb]
  \centering
  \includegraphics[clip, width=0.95\hsize]{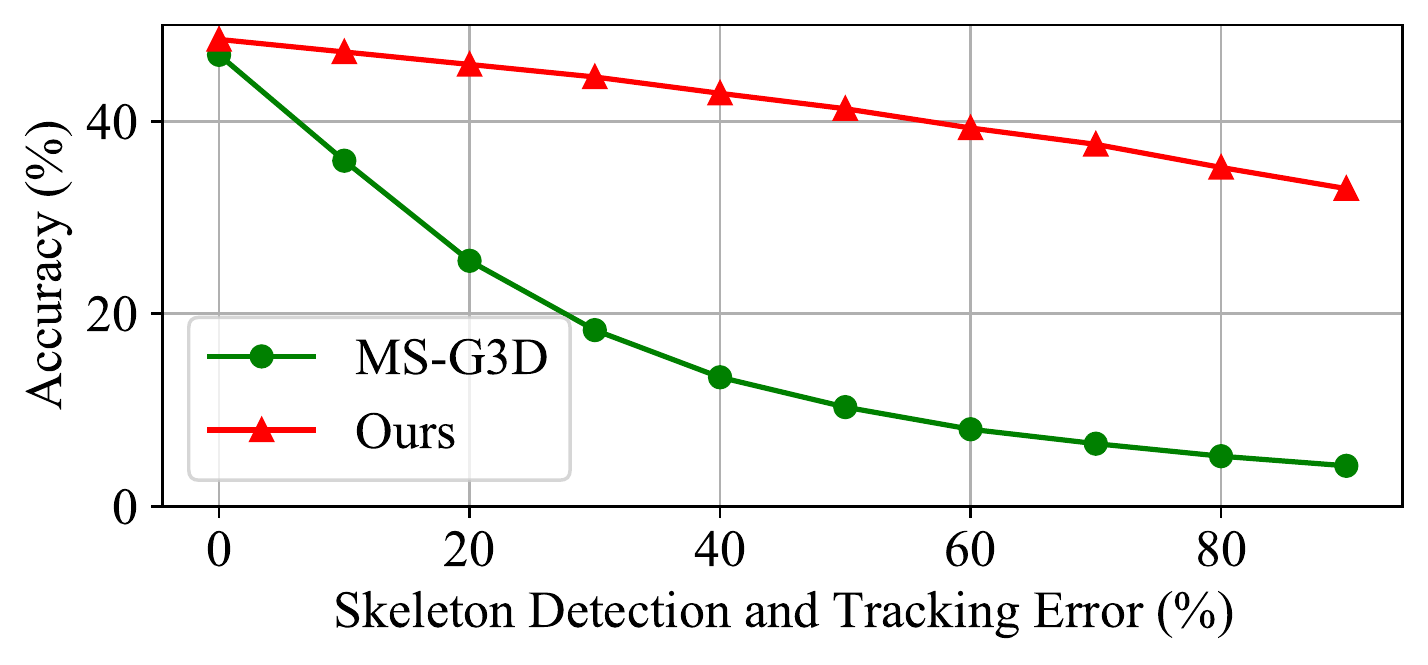}
  \caption{Comparison of the robustness against skeleton detection and tracking errors on the Kinetics-400 dataset. The methods are trained and evaluated using HRNet skeletons for a fair comparison.}
  \label{fig:detection_error}
\end{figure}

The robustness of the proposed method against skeleton detection errors (FPs, FNs, and tracking errors) is compared with that of the MS-G3D~\cite{Liu2020CVPR}, which is the best-performing SoTA method considering both accuracy and runtime metrics, as shown in \cref{tab:kineticsAcc}.
Here, we synthetically generated three types of skeleton detection errors, FPs, FNs, and tracking errors.
The FPs were generated by adding noise sampled from a normal distribution to the keypoint image coordinates.
The FNs were generated by replacing the keypoint image coordinates and the confidence score with $0$ using a certain ratio.
The tracking errors were generated by switching the tracking indices with a certain interval.
Note that the action recognition accuracy of GCN-based methods relies on tracking errors.
On the other hand, the proposed method does not because the proposed network architecture is permutation-invariant for the input keypoints, and the tracking indices are not used.

\cref{fig:detection_error} shows that the performance of the SoTA method (MS-G3D) is highly degraded by adding errors to the inputs.
In contrast, since the performance degradation of the proposed method is relatively small, the proposed method is robust against skeleton detection and tracking errors, which are described as the first robustness limitation in \cref{sec:intro}.

\subsection{Action Recognition Accuracy Comparison with Appearance-based Approaches}

In \cref{tab:other_datasets}, the performance of the proposed method is compared against that of both the SoTA skeleton-based and appearance-based approaches that use RGB and/or optical flow images as inputs.
Here, for a fair comparison with SoTA skeleton-based approaches~\cite{Su2020ECCV,Moon2021CVPR,Duan2022CVPR}, only HRNet skeletons are used as input.
HRNet exhibits a detection performance similar to that of conventional skeleton detection methods employed in these approaches.

The RWF-2000 dataset captures two actions (violence and non-violence) using surveillance cameras in a similar environment.
In the Mimetics experiment, the DNNs are trained with the Kinetics-400 dataset, while the appearances of a person or a background are different from the Kinetics-400 in the evaluation videos.
Hence, the skeleton-based approaches outperform the appearance-based approaches when the RWF-2000 and Mimetics datasets are employed. The opposite occurs when the UCF101 and HMDB51 datasets are employed.
Conclusively, the performance of each of the two approaches depends on the pair of datasets employed.
This result, that the appearance-based approaches are highly biased to background or person appearances, is also mentioned in previous studies~\cite{Choi2019Neurips,Weinzaepfel2021IJCV,Moon2021CVPR}.

The proposed method outperforms the SoTA methods by a certain margin, except for UCF101 and HMDB51 datasets.
In particular, the proposed method outperforms that of SPIL~\cite{Su2020ECCV}, which handles the sequential skeleton data as a point cloud on four violence recognition datasets.

\subsection{Ablation Studies}
\label{sec:ablation}

\noindent \textbf{Effect of the GPB.}
An ablation study of the GPB, which aggregates keypoint features using prior knowledge of the keypoint belongings, instances, or frames, is shown in \cref{tab:gpb_ablation}.
Three models are compared; the model where the GPB is not applied at the two stages (instance-level and frame-level), the model where the GPB is applied only at the first stage, and the proposed model.
Instead of not using the GPB, we concatenate the feature vector from GMPool and each input vector.
It can be observed that the first GPB, which aggregates the features into the instance level, significantly improves the accuracy and speed.
The second GPB mainly improves the runtime.

\noindent \textbf{Effect of Object Keypoints.}
\cref{tab:obj_ablation} shows the results obtained using object categories and eight object contour keypoints as an additional input with the Kinetics-400 dataset.
It can be observed that compared to the accuracy of the skeleton-only input ($41.2\%$), the accuracy of the proposed model is improved when using object categories and four bounding box points ($48.6\%$) as an additional input.
Furthermore, the accuracy is further improved by introducing eight object contour points ($49.2\%$) instead of bounding box points, and both the category and object contour points are informative of the action recognition task.

\noindent \textbf{Ablation Study of the Overall Framework.} 
An ablation study is performed to verify the GPB and MLP Block contributions in \cref{tab:detailed_ablation}.
Also, \cref{tab:detailed_ablation} includes the results of the model replacing our first MLP Block with the GCN-based MS-G3D module~\cite{Liu2020CVPR}.
The simplest baseline (top-left cell) extracts point-wise features and aggregates them using MLPs and GMPool, respectively, similar to PointNet~\cite{Qi2017CVPR}. 
This baseline performs poorly; the GPB yields significant enhancement in accuracy (30.3\% vs. 44.5\%).
Moreover, our method outperforms the version employing the MS-G3D module, which models the temporal information among keypoints (45.7\% vs. 48.5\%).

\subsection{Domain Shift by Introducing Object}
 \label{sec:domain_shift}
An accuracy comparison with and without using object keypoint information is summarized in \cref{tab:dom_shift}.
Here, the models are trained using a synthetically-created Mixamo dataset and evaluated using a real Kinetics dataset to reproduce a challenging, cross-dataset domain shift.
In addition, the proposed method is compared with an appearance-based method~\cite{Carreira2017CVPR} and the SoTA unsupervised and weakly supervised domain adaptation methods~\cite{Chen2019ICCV,Costa2022WACV}.

It can be observed that the accuracy is improved by introducing the proposed object keypoints ($27.6\%$ vs. $28.4\%$), and the variety of actions is expanded without overfitting (the second target-action limitation mentioned in \cref{sec:intro}).
Furthermore, since the proposed method outperforms the appearance-based approaches without any domain adaptation or supervision of the target dataset (Kinetics-400), it is suitable for practical cases when the scene appearance differs between the training and inference phases.

\subsection{Spatio-Temporal Action Localization}
The weakly supervised spatio-temporal action localization accuracy is summarized in \cref{tab:act_loc}.
Since no previous study addressed the task of using only skeletons as an input, the proposed method is compared against appearance-based approaches~\cite{Victor2020CVIU,Cheron2018Neurips,Anurag2020ECCV} and the evaluation protocol~\cite{Anurag2020ECCV} is followed, although the UCF101 is an advantageous dataset for the appearance-based approaches shown in \cref{tab:other_datasets}.

The proposed method without batch-mixing augmentation outperforms SoTA weakly supervised action localization methods with the AP@0.5 metric.
Furthermore, the proposed method outperforms these methods with both AP@0.2 and AP@0.5 metrics by introducing batch-mixing augmentation.
Therefore, as mentioned in the third multi-action limitation in \cref{sec:intro}, the proposed method localizes the actions of each person in each frame.
The qualitative results of the action localization are shown in \cref{fig:teaser} (bottom).

\section{Conclusion}
In this paper, a novel framework with a DNN architecture, Structured Keypoint Pooling, was proposed to address the limitations of conventional skeleton-based action recognition methods.
Time-series keypoints consisting of human skeletons and nonhuman object contours were treated as an input 3D point cloud of the Structured Keypoint Pooling, which sparsely aggregates keypoint features into a cascaded manner based on prior knowledge of the data structure to which the keypoints belong.
A Pooling-Switching Trick, which switches the aggregation target in the phases, and novel data augmentation, which mixes multiple point clouds, were also proposed.
We comprehensively verified the effectiveness against the limitations using several action recognition and localization datasets.
The experimental results demonstrated that the proposed method outperforms SoTA methods regarding both skeleton-based action recognition and spatio-temporal action localization tasks.

{\small
\bibliographystyle{ieee_fullname}
\bibliography{egbib}
}

\end{document}